\documentclass[11pt,a4paper]{article}
\usepackage[hyperref]{acl2018}
\usepackage{times}
\usepackage{latexsym}
\usepackage{url}

\usepackage{arydshln} 
\usepackage{graphicx} 
\usepackage{listings} 
\usepackage{multirow} 
\usepackage{xspace}

\hypersetup{pdftitle={Improving Text-to-SQL Evaluation Methodology},
pdfauthor={Catherine Finegan-Dollak,  Jonathan K. Kummerfeld, et al.},
pdfsubject={Natural Language Processing},
pdfpagelayout=OneColumn}

\newcommand{\tightparagraph}[1]{\paragraph{#1}}
\newcommand{\sql}[1]{\texttt{{\footnotesize #1}}}

\newcommand{\copyWord}{\texttt{COPY}\@\xspace}

\aclfinalcopy 
\def\aclpaperid{1070} 

\title{Improving Text-to-SQL Evaluation Methodology}
\author{
\begin{tabular}{cc}
  \multicolumn{2}{c}{Catherine Finegan-Dollak$^1$\thanks{~~The first two authors contributed equally to this work.} \hspace{1cm} Jonathan K. Kummerfeld$^1$\footnotemark[1] \hspace{1cm} Li Zhang$^1$} \\
  \multicolumn{2}{c}{Karthik Ramanathan$^2$ \hspace{1cm}  Sesh Sadasivam$^1$ \hspace{1cm} Rui Zhang$^3$ \hspace{1cm} Dragomir Radev$^3$} \\
  {\normalfont Computer Science \& Engineering$^1$ and School of Information$^2$} & {\normalfont Department of Computer Science$^3$} \\
  {\normalfont University of Michigan, Ann Arbor} & {\normalfont Yale University} \\
  {\tt \{cfdollak,jkummerf\}@umich.edu} & {\tt dragomir.radev@yale.edu}
\end{tabular}
}

\date{}

\begin{document}
\maketitle

\begin{abstract}

To be informative, an evaluation must measure how well systems generalize to realistic unseen data.
We identify limitations of and propose improvements to current evaluations of text-to-SQL systems.
First, we compare human-generated and automatically generated questions, characterizing properties of queries necessary for real-world applications.
To facilitate evaluation on multiple datasets, we release standardized and improved versions of seven existing datasets and one new text-to-SQL dataset.
Second, we show that the current division of data into training and test sets measures robustness to variations in the way questions are asked, but only partially tests how well systems generalize to new queries; therefore, we propose a complementary dataset split for evaluation of future work.
Finally, we demonstrate how the common practice of anonymizing variables during evaluation removes an important challenge of the task.
Our observations highlight key difficulties, and our methodology enables effective measurement of future development.
\end{abstract}

\section{Introduction}

Effective natural language interfaces to databases (NLIDB) would give lay people access to vast amounts of data stored in relational databases.
This paper identifies key oversights in current evaluation methodology for this task.
In the process, we (1) introduce a new, challenging dataset, (2) standardize and fix many errors in existing datasets, and (3) propose a simple yet effective baseline system.\footnote{Code and data is available at \url{https://github.com/jkkummerfeld/text2sql-data/}}

\begin{figure}
\centering
  \includegraphics[width=1.04\columnwidth]{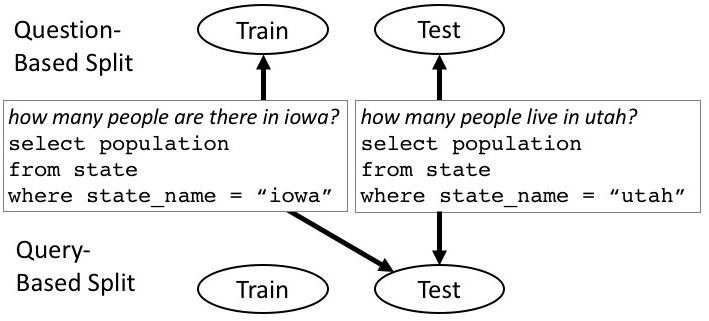}
  \vspace{-3mm}
  \caption{\label{fig:split}
  Traditional question-based splits allow queries to appear in both train and test.
  Our query-based split ensures each query is in only one.}
\end{figure}

First, we consider query complexity, showing that human-written questions require more complex queries than automatically generated ones.
To illustrate this challenge, we introduce \textit{Advising}, a dataset of questions from university students about courses that lead to particularly complex queries.

Second, we identify an issue in the way examples are divided into training and test sets.
The standard approach, shown at the top of Fig.~\ref{fig:split}, divides examples based on the text of each \textit{question}.
As a result, many of the queries in the test set are seen in training, albeit with different entity names and with the question phrased differently.
This means metrics are mainly measuring robustness to the way a set of known SQL queries can be expressed in English---still a difficult problem, but not a complete test of ability to compose new queries in a familiar domain.
We introduce a template-based slot-filling baseline that cannot generalize to new queries, and yet is competitive with prior work on multiple datasets.
To measure robustness to new queries, we propose splitting based on the SQL \textit{query}.
We show that state-of-the-art systems with excellent performance on traditional question-based splits struggle on query-based splits.
We also consider the common practice of variable anonymization, which removes a challenging form of  ambiguity from the task.
In the process, we apply extensive effort to standardize datasets and fix a range of errors.

Previous NLIDB work has led to impressive systems, but current evaluations provide an incomplete picture of their strengths and weaknesses.
In this paper, we provide new and improved data, a new baseline, and guidelines that complement existing metrics, supporting future work.

\section{Related Work}\label{sec:related}

The task of generating SQL representations from English questions has been studied in the NLP and DB communities since the 1970s \cite{Androutsopoulos1995}.
Our observations about evaluation methodology apply broadly to the systems cited below.

Within the DB community, systems commonly use pattern matching, grammar-based techniques, or intermediate representations of the query \cite{PazosR.2013}.
Recent work has explored incorporating user feedback to improve accuracy \cite{Li2014}.
Unfortunately, none of these systems are publicly available, and many rely on domain-specific resources.

In the NLP community, there has been extensive work on semantic parsing to logical representations that query a knowledge base \cite{Zettlemoyer2005,Liang2011,Beltagy2014,Berant2014}, while work on mapping to SQL has recently increased \cite{Yih2015,Iyer2017,Zhong2017}.
One of the earliest statistical models for mapping text to SQL was the PRECISE system \cite{Popescu2003,Popescu2004}, which achieved high precision on queries that met constraints linking tokens and database values, attributes, and relations, but did not attempt to generate SQL for questions outside this class.
Later work considered generating queries based on relations extracted by a syntactic parser \cite{Giordani2012} and applying techniques from logical parsing research \cite{Poon2013}.
However, none of these earlier systems are publicly available, and some required extensive engineering effort for each domain, such as the lexicon used by PRECISE.

More recent work has produced general purpose systems that are competitive with previous results and are also available, such as \newcite{Iyer2017}.
We also adapt a logical form parser with a sequence to tree approach that makes very few assumptions about the output structure \cite{Dong2016}.

One challenge for applying neural models to this task is annotating large enough datasets of question-query pairs.
Recent work \cite{Cai2017,Zhong2017} has automatically generated large datasets using templates to form random queries and corresponding natural-language-like questions, and then having humans rephrase the question into English.
Another option is to use feedback-based learning, where the system alternates between training and making predictions, which a user rates as correct or not \cite{Iyer2017}.
Other work seeks to avoid the data bottleneck by using end-to-end approaches \cite{Yin2016,Neelakantan2017}, which we do not consider here.
One key contribution of this paper is standardization of a range of datasets, to help address the challenge of limited data resources.

\section{Data}\label{sec:data}
For our analysis, we study a range of text-to-SQL datasets, standardizing them to have a consistent SQL style.

\tightparagraph{ATIS} \cite{Price1990,Dahl1994} User questions for a flight-booking task, manually annotated.
We use the modified SQL from \newcite{Iyer2017}, which follows the data split from the logical form version \cite{Zettlemoyer2007}.

\tightparagraph{GeoQuery} \cite{Zelle1996} User questions about US geography, manually annotated with Prolog.
We use the SQL version \cite{Popescu2003,Giordani2012,Iyer2017}, which follows the logical form data split \cite{Zettlemoyer2005}.

\tightparagraph{Restaurants} \cite{Tang2000,Popescu2003} User questions about restaurants, their food types, and locations. 

\tightparagraph{Scholar} \cite{Iyer2017} User questions about academic publications, with automatically generated SQL that was checked by asking the user if the output was correct.

\tightparagraph{Academic} \cite{Li2014} Questions about the Microsoft Academic Search (MAS) database, derived by enumerating every logical query that could be expressed using the search page of the MAS website and writing sentences to match them.
The domain is similar to that of Scholar, but their schemas differ. 

\tightparagraph{Yelp and IMDB} \cite{Yaghmazadeh2017} Questions about the Yelp website and the Internet Movie Database, collected from colleagues of the authors who knew the type of information in each database, but not their schemas.

\tightparagraph{WikiSQL} \cite{Zhong2017} A large collection of automatically generated questions about individual tables from Wikipedia, paraphrased by crowd workers to be fluent English.

\tightparagraph{Advising (This Work)} Our dataset of questions over a database of course information at the University of Michigan, but with fictional student records.
Some questions were collected from the EECS department Facebook page and others were written by CS students with knowledge of the database who were instructed to write questions they might ask in an academic advising appointment. 

The authors manually labeled the initial set of questions with SQL.
To ensure high quality, at least two annotators scored each question-query pair on a two-point scale for accuracy---did the query generate an accurate answer to the question?---and a three-point scale for helpfulness---did the answer provide the information the asker was probably seeking?
Cases with low scores were fixed or removed from the dataset. 

We collected paraphrases using \newcite{Jiang2017}'s method, with manual inspection to ensure accuracy.
For a given sentence, this produced paraphrases with the same named entities (e.g. course number EECS 123).
To add variation, we annotated entities in the questions and queries with their types---such as course name, department, or instructor---and substituted randomly-selected values of each type into each paraphrase and its corresponding query.
This combination of paraphrasing and entity replacement means an original question of ``For next semester, who is teaching EECS 123?" can give rise to ``Who teaches MATH 456 next semester?" as well as ``Who's the professor for next semester's CHEM 789?"

\subsection{SQL Canonicalization}\label{subsec:canonicalization}

SQL writing style varies.
To enable consistent training and evaluation across datasets, we canonicalized the queries:
(1) we alphabetically ordered fields in \sql{SELECT}, tables in \sql{FROM}, and constraints in \sql{WHERE};
(2) we standardized table aliases in the form \sql{<TABLE\_NAME>alias<N>} for the \sql{N}th use of the same table in one query; and
(3) we standardized capitalization and spaces between symbols. 
We confirmed these changes do not alter the meaning of the queries via unit tests of the canonicalization code and manual inspection of the output.
We also manually fixed some errors, such as ambiguous mixing of \sql{AND} and \sql{OR} (30 ATIS queries).

\subsection{Variable Annotation}\label{subsec:variables}
Existing SQL datasets do not explicitly identify which words in the question are used in the SQL query.
Automatic methods to identify these variables, as used in prior work, do not account for ambiguities, such as words that could be either a city or an airport.
To provide accurate anonymization, we annotated query variables using a combination of automatic and manual processing.

Our automatic process extracted terms from each side of comparison operations in SQL: one side contains quoted text or numbers, and the other provides a type for those literals.
Often quoted text in the query is a direct copy from the question, while in some cases we constructed dictionaries to map common acronyms, like \textit{american airlines}--\sql{AA}, and times, like \textit{2pm}--\sql{1400}.
The process flagged cases with ambiguous mappings, which we then manually processed.
Often these were mistakes, which we corrected, such as missing constraints (e.g., \textit{papers in 2015} with no date limit in the query), extra constraints (e.g., limiting to a single airline despite no mention in the question), inaccurate constraints (e.g., \textit{more than 5} as \sql{$>$ 4}), and inconsistent use of \textit{this year} to mean different years in different queries.

\subsection{Query Deduplication} \label{subsec:manual-repair}
\begin{table}
\small
\begin{center}
\begin{tabular}{l c c}
\hline
& Sets Identified & Affected Queries\\
ATIS & 141 & 380\\
GeoQuery & 17 & 39\\
Scholar & 60 & 152\\
\hline
\end{tabular}
\caption{Manually identified duplicate queries (different SQL for equivalent questions).}
\label{table:duplicates}
\end{center}
\end{table}

Three of the datasets had many duplicate queries (i.e., semantically equivalent questions with different SQL).
To avoid this spurious ambiguity we manually grouped the data into sets of equivalent questions (Table \ref{table:duplicates}).
A second person manually inspected every set and ran the queries.
Where multiple queries are valid, we kept them all, though only used the first for the rest of this work.

\begin{table*}
\small
\setlength{\tabcolsep}{3pt}
\begin{center}
\begin{tabular}{lc c c@{\hskip16pt}c c@{\hskip16pt} c | c c@{\hskip16pt}c c@{\hskip16pt}c c@{\hskip16pt}c c } 
\hline
&\multicolumn{6}{c|}{Redundancy Measures}
&\multicolumn{8}{c}{Complexity Measures} \\
&\multicolumn{1}{c}{}
&\multicolumn{1}{c}{Unique}
&\multicolumn{1}{c}{}
&\multicolumn{2}{l}{\hspace{3mm}Queries}
&\multicolumn{1}{c|}{}
&\multicolumn{2}{l}{\hspace{2mm}Tables} 
&\multicolumn{2}{l}{\hspace{-3mm}Unique tables} 
&\multicolumn{2}{l}{SELECTs} 
&\multicolumn{2}{c}{Nesting}
\\
&\multicolumn{1}{c}{Question}
&\multicolumn{1}{c}{query}
&\multicolumn{1}{c}{}
&\multicolumn{2}{l}{\hspace{3mm}/ pattern}
&\multicolumn{1}{c|}{Pattern}
&\multicolumn{2}{l}{\hspace{2mm}/ query} 
&\multicolumn{2}{l}{\hspace{1mm}/ query} 
&\multicolumn{2}{l}{\hspace{2mm}/ query} 
&\multicolumn{2}{c}{Depth}  
\\
& count 
& count 
& [1]/[2] 
& $\mu$ & Max 
& count 
& $\mu$ & Max 
& $\mu$ & Max 
& $\mu$ & Max 
& $\mu$ & Max  
\\
\hline
\hline
Advising & 4570 & 211 & 21.7 & 20.3 & 90 & 174 
& 3.2 & 9 & 3.0 & 9 & 1.23 & 6 & 1.18 & 4 \\
ATIS & 5280 & 947 & 5.6 & 7.0 & 870 & 751 
& 6.4 & 32 & 3.8 & 12 & 1.79 & 8 & 1.39 & 8 \\
GeoQuery & 877 & 246 & 3.6 & 8.9 & 327 & 98
& 1.4 & 5 & 1.1 & 4 & 1.77 & 8 & 2.03 & 7 \\
Restaurants & 378 & 23 & 16.4 & 22.2 & 81 & 17
& 2.6 & 5 & 2.3 & 4 & 1.17 & 2 & 1.17 & 2 \\
Scholar & 817 & 193 & 4.2 & 5.6 & 71 & 146 
& 3.3 & 6 & 3.2 & 6 & 1.02 & 2 & 1.02 & 2 \\
\hdashline
Academic & 196 & 185 & 1.1 & 2.1 & 12 & 92 
& 3.2 & 10 & 3 & 6 & 1.04 & 3 & 1.04 & 2 \\
IMDB & 131 & 89 & 1.5 & 2.5 & 21 & 52 
& 1.9 & 5 & 1.9 & 5 & 1.01 & 2 & 1.01 & 2 \\
Yelp & 128 & 110 &1.2& 1.4 & 11 & 89 
& 2.2 & 4 & 2 & 4 & 1 & 1 & 1 & 1 \\
\hdashline
WikiSQL & 80,654 & 77,840 & 1.0 & 165.3 & 42,816 & 488 
& 1 & 1 & 1 & 1 & 1 & 1 & 1 & 1 \\
\hline

\end{tabular}
\caption{Descriptive statistics for text-to-SQL datasets. Datasets in the first group are human-generated from the NLP community, in the second are human-generated from the DB community, and in the third are automatically-generated. [1]/[2] is Question count / Unique query count.}
\label{table:dataset_complexities}
\end{center}
\end{table*}

\section{Evaluating on Multiple Datasets Is Necessary}\label{sec:complexity_measures}
For evaluation to be informative it must use data that is representative of real-world queries.
If datasets have biases, robust comparisons of models will require evaluation on multiple datasets.
For example, some datasets, such as ATIS and Advising, were collected from users and are task-oriented, while others, such as WikiSQL, were produced by automatically generating queries and engaging people to express the query in language.
If these two types of datasets differ systematically, evaluation on one may not reflect performance on the other.
In this section, we provide descriptive statistics aimed at understanding how several datasets differ, especially with respect to query redundancy and complexity.

\subsection{Measures}
We consider a range of measures that capture different aspects of data complexity and diversity:

\tightparagraph{Question / Unique Query Counts} We measure dataset size and how many distinct queries there are when variables are anonymized. We also present the mean number of questions per unique query; a larger mean indicates greater redundancy.

\begin{figure}
  \scriptsize
  \noindent
  \sql{SELECT \textit{<table-alias>}.\textit{<field>}\\[2pt]
  FROM \textit{<table>} AS \textit{<table-alias>}\\[2pt]
  WHERE \textit{<table-alias>}.\textit{<field>} = \textit{<literal>}} \\[-2pt]
  
  \noindent
  \sql{SELECT RIVERalias0.RIVER\_NAME \\[2pt]
  FROM RIVER AS RIVERalias0 \\[2pt]
  WHERE RIVERalias0.TRAVERSE = "florida";} \\[-2pt]
  
  \noindent
  \sql{SELECT CITYalias0.CITY\_NAME \\[2pt]
  FROM CITY AS CITYalias0 \\[2pt]
  WHERE CITYalias0.STATE\_NAME = "alabama";}
  
  \caption{\label{fig:patterns}
  An SQL pattern and example queries.
  }
\end{figure}

\tightparagraph{SQL Patterns}
Complexity can be described as the answer to the question, ``How many query-form patterns would be required to generate this dataset?"
Fig.~\ref{fig:patterns} shows an example of a pattern, which essentially abstracts away from the specific table and field names.
Some datasets were generated from patterns similar to these, including WikiSQL and \newcite{Cai2017}.
This enables the generation of large numbers of queries, but limits the variation between them to only that encompassed by their patterns.
We count the number of patterns needed to cover the full dataset, where larger numbers indicate  greater diversity.
We also report mean queries per pattern; here, larger numbers indicate greater redundancy, showing that many queries fit the same mold. 

\tightparagraph{Counting Tables}
We consider the total number of tables and the number of unique tables mentioned in a query.
These numbers differ in the event of self-joins. 
In both cases, higher values imply greater complexity.

\tightparagraph{Nesting}
A query with nested subqueries may be more complex than one without nesting.
We count SELECT statements within each query to determine the number of sub-queries.
We also report the depth of query nesting.
In both cases, higher values imply greater complexity.


\subsection{Analysis}
The statistics in Table~\ref{table:dataset_complexities} show several patterns.

First, dataset size is not the best indicator of dataset diversity. 
Although WikiSQL contains fifteen times as many question-query pairs as ATIS, ATIS contains significantly more patterns than WikiSQL; moreover, WikiSQL's queries are dominated by one pattern that is more than half of the dataset (\sql{SELECT col AS result FROM table WHERE col = value}). 
The small, hand-curated datasets developed by the database community---Academic, IMDB, and Yelp---have noticeably less redundancy as measured by questions per unique query and queries per pattern than the datasets the NLP community typically evaluates on. 

Second, human-generated datasets exhibit greater complexity than automatically generated data. 
All of the human-generated datasets except Yelp demonstrate at least some nesting. The average query from any of the human-generated datasets joins more than one table. 

In particular, task-oriented datasets require joins and nesting. ATIS and Advising, which were developed with air-travel and student-advising tasks in mind, respectively, both score in the top three for multiple complexity scores.

To accurately predict performance on human-generated or task-oriented questions, it is thus necessary to evaluate on datasets that test the ability to handle nesting and joins. 
Training and testing NLP systems, particularly deep learning-based methods, benefits from large datasets. 
However, at present, the largest dataset available does not provide the desired complexity.

\tightparagraph{Takeaway:} Evaluate on multiple datasets, some with nesting and joins, to provide a thorough picture of a system's strengths and weaknesses.  

\section{Current Data Splits Only Partially Probe Generalizability}\label{sec:experiments}

It is standard best practice in machine learning to divide data into disjoint training, development, and test sets.
Otherwise, evaluation on the test set will not accurately measure how well a model generalizes to new examples.
The standard splits of GeoQuery, ATIS, and Scholar treat each pair of a natural language question and its SQL query as a single item.
Thus, as long as each question-query pair appears in only one set, the test set is not tainted with training data.
We call this a question-based data split. 

However, many English questions may correspond to the same SQL query. If at least one copy of every SQL query appears in training, then the task evaluated is classification, not true semantic parsing, of the English questions. We can increase the number of distinct SQL queries by varying what entities our questions ask about; the queries for \textit{what states border Texas} and \textit{what states border Massachusetts} are not identical. Adding this variation changes the task from pure classification to classification plus slot-filling. Does this provide a true evaluation of the trained model's performance on unseen inputs? 

It depends on what we wish to evaluate.
If we want a system that answers questions within a particular domain, and we have a dataset that we are confident covers everything a user might want to know about that domain, then evaluating on the traditional question-based split tells us whether the system is robust to variation in how a request is expressed.
But compositionality is an essential part of language, and a system that has trained on \textit{What courses does Professor Smith teach?} and \textit{What courses meet on Fridays?} should be prepared for \textit{What courses that Professor Smith teaches meet on Fridays?} Evaluation on the question split does not tell us about a model's generalizable knowledge of SQL, or even its generalizable knowledge within the present domain. 

To evaluate the latter, we propose a complementary new division, where no SQL query is allowed to appear in more than one set; we call this the \textit{query split}. To generate a query split, we substitute variables for entities in each query in the dataset, as described in \S~\ref{subsec:variables}. Queries that are identical when thus anonymized are treated as a single query and randomly assigned---with all their accompanying questions---to train, dev, or test. We include the original question split and the new query split labeling for the new Advising dataset, as well as ATIS, GeoQuery, and Scholar. For the much smaller Academic, IMDB, Restaurant, and Yelp datasets, we include question- and query- based buckets for cross validation.  

\subsection{Systems}

Recently, a great deal of work has used variations on the seq2seq model.
We compare performance of a basic seq2seq model \cite{Sutskever2014}, and seq2seq with attention over the input \cite{Bahdanau2014}, implemented with TensorFlow seq2seq \cite{Britz2017}.
We also extend that model to include an attention-based copying option, similar to \newcite{Jia2016}. 
Our output vocabulary for the decoder includes a special token, \copyWord.
If \copyWord has the highest probability at step $t$, we replace it with the input token with the max of the normalized attention scores.
Our loss function is the sum of two terms:
first, the categorical cross entropy for the model's probability distribution over the output vocabulary tokens; and
second, the loss for word copying.
When the correct output token is \copyWord, the second loss term is the categorical cross entropy of the distribution of attention scores at time $t$.
Otherwise it is zero.

For comparison, we include systems from two recent papers.
\newcite{Dong2016} used an attention-based seq2tree model for semantic parsing of logical forms; we apply their code here to SQL datasets.
\newcite{Iyer2017} use a seq2seq model with automatic dataset expansion through paraphrasing and SQL templates.\footnote{
We enable \newcite{Iyer2017}'s paraphrasing data augmentation, but not their template-based augmentation because templates do not exist for most of the datasets (though they also found it did not significantly improve performance).
Note, on ATIS and Geo their evaluation assumed no ambiguity in entity identification, which is equivalent to our Oracle Entities condition (\S \ref{subsec:oracle_entity}).
}

We could not find publicly available code for the non-neural text-to-SQL systems discussed in Section~\ref{sec:related}.
Also, most of those approaches require development of specialized grammars or templates for each new dataset they are applied to, so we do not compare such systems. 

\begin{figure}
    \centering
    \includegraphics[width=\linewidth, trim={7cm 9.4cm 7.5cm 10.8cm}, clip]{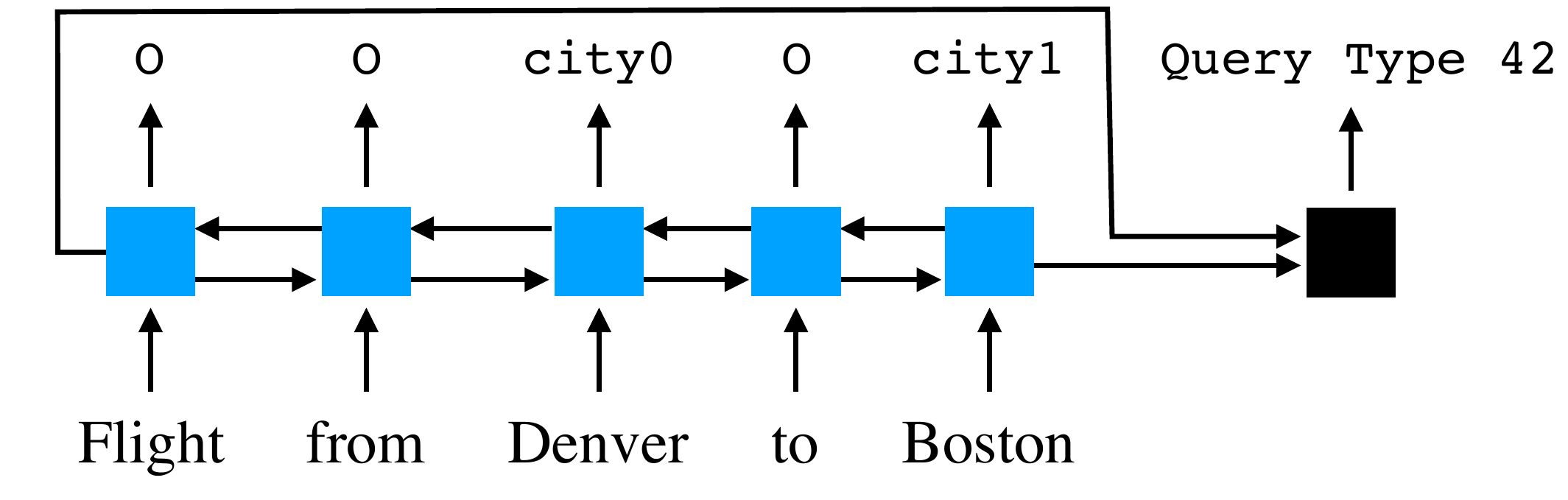}
    \caption{\label{fig:baseline}
    Baseline: blue boxes are LSTM cells and the black box is a feed-forward network.
    Outputs are the query template to use (right) and which tokens to fill it with (left).}
\end{figure}

\begin{table*}
\small
\setlength{\tabcolsep}{5pt}
\begin{center}
\begin{tabular}{l rr c rr c rr c rr c rr | c rr c rr c rr}
  \hline
  & \multicolumn{2}{c}{Advising}
  & & \multicolumn{2}{c}{ATIS}
  & \multicolumn{3}{c}{GeoQuery}
  & \multicolumn{4}{c}{Restaurants}
  & \multicolumn{2}{c}{Scholar}
  & \multicolumn{3}{c}{Academic}
  & & \multicolumn{2}{c}{IMDB}
  & & \multicolumn{2}{c}{Yelp}
  \\
  Model
  & ? & \sql{Q}
  & & ? & \sql{Q}
  & & ? & \sql{Q}
  & & ? & \sql{Q}
  & & ? & \sql{Q}
  & & ? & \sql{Q}
  & & ? & \sql{Q}
  & & ? & \sql{Q}\\
  \hline
  \hline
  \multicolumn{24}{c}{No Variable Anonymization} \\
    Baseline
    & \textbf{80} &  0 
    & & 46 &  0 
    & & 57 &  0 
    & & 95 &  0 
    & & 52 &  0 
    & &  0 &  0 
    & &  0 &  0 
    & &  1 &  0 
    \\
    seq2seq
    &  6 &  0 
    & &  8 &  0 
    & & 27 &  7 
    & & 47 &  0 
    & & 19 &  0 
    & &  6 &  7 
    & &  1 &  0 
    & &  0 &  0 
    \\
    \hspace{2mm}+ Attention
    & 29 &  0 
    & & 46 & 18 
    & & 63 & 21 
    & & \textbf{100} &  2 
    & & 33 &  0 
    & & 71 & 64 
    & &  7 &  3 
    & &  2 &  2 
    \\
    \hspace{2mm}+ Copying
    & 70 &  0 
    & & \textbf{51} & \textbf{32} 
    & & \textbf{71} & 20 
    & & \textbf{100} &  4 
    & & \textbf{59} &  5 
    & & \textbf{81} & \textbf{74} 
    & & \textbf{26} & \textbf{9} 
    & & \textbf{12} &  4 
    \\
    D\&L seq2tree
    & 46 &  \textbf{2} 
    & & 46 & 23 
    & & 62 & 31 
    & & \textbf{100} & \textbf{11} 
    & & 44 &  \textbf{6} 
    & & 63 & 54 
    & &  6 &  2 
    & &  1 &  2 
    \\
    Iyer et al.
    & 41 &  1 
    & & 45 & 17 
    & & 66 & \textbf{40} 
    & & \textbf{100} &  8 
    & & 44 &  3 
    & & 76 & 70 
    & & 10 &  4 
    & &  6 & \textbf{6} 
    \\
  \hline
  \multicolumn{24}{c}{With Oracle Entities} \\
 Baseline
    & 89 &  0 
    & & 56 &  0 
    & & 56 &  0 
    & & 95 &  0 
    & & 66 &  0 
    & &  0 &  0 
    & &  7 &  0 
    & &  8 &  0 
    \\
    seq2seq
    & 21 &  0 
    & & 14 &  0 
    & & 49 &  14 
    & &	71	&	6	
    & & 23 &  0 
    & & 10 &  9 
    & & 6	&	0	
    & &	12	&	9 
    \\
    \hspace{2mm}+ Attention
    & 88 &  0 
    & & 57 & 23 
    & & 73 & 31 
    & &	100	&	32
    & & 71 &  4 
    & &	77	& 74 
    & &	44	&	17	
    & &	33	&28	
    \\
    D\&L seq2tree
    & 88 &  8 
    & & 56 & 34 
    & & 68 & 23 
    & & 100	&	21	
    & & 68 &  6 
    & &	65	&	61	
    & &	36	&	10	
    & &	26	&	23
    \\
    Iyer et al.
    & 88 &  6 
    & & 58 & 32 
    & & 71 & 49 
    & &	100	&	33	
    & & 71 &  1 
    & &	77	&	75	
    & &	52	&	24	
    & &	44	&	32  
    \\
  \hline
    Baseline-Oracle
    & 100 &  0 
    & & 69 &  0 
    & & 78 &  0 
    & & 100 &  0 
    & & 84 &  0 
    & & 11 &  0 
    & & 47 &  0 
    & & 25 &  0 
    \\
  \hline
\end{tabular}
\caption{Accuracy of neural text-to-SQL systems on English question splits (`?' columns) and SQL query splits (`\sql{Q}' columns).
The vertical line separates datasets from the NLP (left) and DB (right) communities.
Results for \newcite{Iyer2017} are slightly lower here than in the original paper because we evaluate on SQL output, not the database response.
}
\label{table:sql_results}
\end{center}
\end{table*}

\subsection{New Template Baseline}

In addition to the seq2seq models, we develop a new baseline system for text-to-SQL parsing which exploits repetitiveness in data. 
First, we automatically generate SQL templates from the training set.
The system then makes two predictions: (1) which template to use, and (2) which words in the sentence should fill slots in the template.
This system is not able to generalize beyond the queries in the training set, so it will fail completely on the new query-split data setting.

Fig.~\ref{fig:baseline} presents the overall architecture, which we implemented in DyNet \cite{dynet}.
A bidirectional LSTM provides a prediction for each word, either \texttt{O} if the word is not used in the final query, or a symbol such as \sql{city1} to indicate that it fills a slot.
The hidden states of the LSTM at each end of the sentence are passed through a small feed-forward network to determine the SQL template to use.
This architecture is simple and enables a joint choice of the tags and the template, though we do not explicitly enforce agreement.

To train the model, we automatically construct a set of templates and slots.
Slots are determined based on the variables in the dataset, with each SQL variable that is explicitly given in the question becoming a slot.
We can construct these templates because our new version of the data explicitly defines all variables, their values, and where they appear in both question and query.

For completeness, we also report on an oracle version of the template-based system (performance if it always chose the correct template from the train set and filled all slots correctly). 

\subsection{Oracle Entity Condition}\label{subsec:oracle_entity}
Some systems, such as Dong and Lapata's model, are explicitly designed to work on anonymized data (i.e., data where entity names are replaced with a variable indicating their type). 
Others, such as attention-based copying models, treat identification of entities as an inextricable component of the text-to-SQL task.
We therefore describe results on both the actual datasets with entities in place and a version anonymized using the variables described in \S~\ref{subsec:variables}. 
We refer to the latter as the oracle entity condition.

\subsection{Results and Analysis}

We hypothesized that even a system unable to generalize can achieve good performance on question-based splits of datasets, and the results  in Table~\ref{table:sql_results} substantiate that for the NLP community's datasets. 
The template-based, slot-filling baseline was competitive with state-of-the-art systems for question split on the four datasets from the NLP community. 
The template-based oracle performance indicates that for these datasets anywhere from 70-100\% accuracy on question-based split could be obtained by selecting a template from the training set and filling in the right slots.

For the three datasets developed by the databases community, the effect of question-query split is far less pronounced.
The small sizes of these datasets cannot account for the difference, since even the oracle baseline did not have much success on these question splits, and since the baseline was able to handle the small Restaurants dataset. 
Looking back at Section~\ref{sec:complexity_measures}, however, we see that these are the datasets with the least redundancy in Table~\ref{table:dataset_complexities}. 
Because their question:unique-query ratios are nearly 1:1, the question splits and query splits of these datasets were quite similar. 

Reducing redundancy does not improve performance on query split, though; at most, it reduces the difference between performance on the two splits.
IMDB and Yelp both show weak results on query split despite their low redundancy. 
Experiments on a non-redundant version of query split for Advising, ATIS, GeoQuery, and Restaurant that contained only one question for each query confirmed this: in each case, accuracy remained the same or declined relative to regular query split. 

Having ruled out redundancy as a cause for the exceptional performance on Academic's query split, we suspect the simplicity of its questions and the compositionality of its queries may be responsible. Every question in the dataset begins \textit{return me} followed by a phrase indicating the desired field, optionally followed by one or more constraints; for instance, \textit{return me the papers by `author\_name0'} and \textit{return me the papers by `author\_name0' on journal\_name0.} 

None of this, of course, is to suggest that question-based split is an easy problem, even on the NLP community's datasets. 
Except for the Advising and Restaurants datasets, even the oracle version of the template-based system is far from perfect.
Access to oracle entities helps performance of non-copying systems substantially, as we would expect.
Entity matching is thus a non-trivial component of the task. 

But the query-based split is certainly more difficult than the question-based split.
Across datasets and systems, performance suffered on query split. Access to oracle entities did not remove this effect. 

Many of the seq2seq models do show some ability to generalize, though. Unlike the template-based baseline, many were able to eek out some performance on query split. 

On question split, ATIS is the most difficult of the NLP datasets, yet on query split, it is among the easiest. 
To understand this apparent contradiction, we must consider what kinds of mistakes systems make and the contexts in which they appear. We therefore analyze the output of the attention-based-copying model in greater detail. 

\begin{table*}
\small
\begin{center}
\begin{tabular}{ll|cc|cc|cc|cc} 
\hline
& &\multicolumn{2}{c|}{Advising} & \multicolumn{2}{c|}{ATIS} 
& \multicolumn{2}{c|}{GeoQuery} & \multicolumn{2}{c}{Scholar} \\
& & Question & Query
 & Question & Query
  & Question & Query
   & Question & Query\\
\hline
\hline
\multirow{2}{*}{Correct}&Count&	369	&	5	&	227	&	111	&	191	&	56	&	129	&	17	\\
&$\mu$ Length&	83.8	&	165.8	&	55.1	&	69.2	&	19.6	&	21.5	&	38.0	&	30.2	\\
\hline																
Entity &Count &	10	&	0	&	1	&	6	&	5	&	0	&	5	&	0	\\
problem   &$\mu$ Length &	111.8	&	N/A	&	28.0	&	71.3	&	17.2	&	N/A	&	42.6	&	N/A	\\
\hline																
Different &Count&	43	&	675	&	94	&	68	&	53	&	84	&	40	&	94	\\
template &$\mu$ Length&	69.8	&	68.4	&	85.8	&	72.1	&	25.6	&	18.0	&	43.9	&	39.8	\\
\hline																
No template &Count&	79	&	25	&	122	&	162	&	30	&	42	&	44	&	204	\\
match    &$\mu$ Length&	88.8	&	90.5	&	113.8	&	92.2	&	29.7	&	25.0	&	42.1	&	41.6	\\
\hline

\end{tabular}
\caption{Types of errors by the attention-based copying model for question and query splits, with (Count)s of queries in each category, and the ($\mu$ Length) of gold queries in the category.}
\label{table:errors}
\end{center}
\end{table*}

We categorize each output as shown in column one of Table \ref{table:errors}. The ``Correct" category is self-explanatory. ``Entity problem only" means that the query would have been correct but for a mistake in one or more entity names.
``Different template" means that the system output was the same as another query from the dataset but for the entity names; however, it did not match the correct query for this question. 
``No template match" contains both the most mundane and the most interesting errors. Here, the system output a query that is not copied from training data. Sometimes, this is a simple error, such as inserting an extra comma in the \sql{WHERE} clause. Other times, it is recombining segments of queries it has seen into new queries. This is necessary but not sufficient model behavior in order to do well on the query split. 
In at least one case, this category includes a semantically equivalent query marked as incorrect by the exact-match accuracy metric.\footnote{For the question \textit{which of the states bordering pennsylvania has the largest population}, the gold query ranked the options by population and kept the top result, while the system output used a subquery to find the max population then selected states that had that population.}
Table \ref{table:errors} shows the number of examples from the test set that fell into each category, as well as the mean length of gold queries (``length") for each category.

Short queries are easier than long ones in the question-based condition. 
In most cases, length in ``correct" is shorter than length in either ``different template" or ``no template match" categories.

In addition, for short queries, the model seems to prefer to copy a query it has seen before; for longer ones, it generates a new query. 
In every case but one, mean length in ``different template" is less than in ``No template match." 

Interestingly, in ATIS and GeoQuery, where the model performs tolerably well on query split, the length for correct queries in query split is higher than the length for correct queries from the question split. Since, as noted above, recombination of template pieces (as we see in ``no template match") is a necessary step for success on query split, it may be that longer queries have a higher probability of recombination, and therefore a better chance of being correct in query split. The data from Scholar does not support this position; however, note that only 17 queries were correct in Scholar query split, suggesting caution in making generalizations from this set.


These results also seem to indicate that our copying mechanism effectively deals with entity identification.
Across all datasets, we see only a small number of entity-problem-only examples. 
However, comparing the rows from Table~\ref{table:sql_results} for seq2seq+Copy at the top and seq2seq+Attention in the oracle entities condition, it is clear that having oracle entities provides a useful signal, with consistent gains in performance.

\tightparagraph{Takeaways:} Evaluate on both question-based and query-based dataset splits.
Additionally, variable anonymization noticeably decreases the difficulty of the task; thus, thorough evaluations should include results on datasets without anonymization. 

\subsection{Logic Variants}\label{subsec:logic}

%

To see if our observations on query and question split performance apply beyond SQL, we also considered the logical form annotations for ATIS and GeoQuery \cite{Zettlemoyer2005,Zettlemoyer2007}.
We retrained \newcite{Jia2016}'s baseline and full system.
Interestingly, we founnd limited impact on performance, measured with either logical forms or denotations.
To understand why, we inspected the logical form datasets.
In both ATIS and GeoQuery, the logical form version has a larger set of queries after variable identification.
This seems to be because the logic abstracts away from the surface form less than SQL does.
For example, these questions have the same SQL in our data, but different logical forms:

\noindent
{\small
 \textit{what state has the largest capital} \\[-4pt]
 {\tt\scriptsize (A, (state(A), loc(B, A), largest(B, capital(B))))} \\[-4pt]
 \textit{which state 's capital city is the largest} \\[-4pt]
 {\tt\scriptsize (A, largest(B, (state(A), capital(A, B), city(B))))} \\
\sql{SELECT CITYalias0.STATE\_NAME} \\[-4pt]
\sql{FROM CITY AS CITYalias0} \\[-4pt]
\sql{WHERE CITYalias0.POPULATION = (} \\[-4pt]
\phantom{\sql{AA}}\sql{SELECT MAX( CITYalias1.POPULATION )} \\[-4pt]
\phantom{\sql{AA}}\sql{FROM CITY AS CITYalias1 ,} \\[-4pt]
\phantom{\sql{AAFROMA}}\sql{STATE AS STATEalias0} \\[-4pt]
\phantom{\sql{AA}}\sql{WHERE STATEalias0.CAPITAL =}\\[-4pt]
\phantom{\sql{AAWHEREB}}\sql{CITYalias1.CITY\_NAME ) ;}
}

Other examples include variation in the logical form between sentences with \textit{largest} and \textit{largest population} even though the associated dataset only has population figures for cities (not area or any other measure of size).
Similarly in ATIS, the logical form will add \sql{(flight \$0)} if the question mentions flights explicitly, making these two queries different, even though they convey the same user intent:

\noindent
{\small \it
what flights do you have from bwi to sfo \\[-4pt]
i need a reservation from bwi to sfo
}

By being closer to a syntactic representation, the queries end up being more compositional, which encourages the model to learn more compositionality than the SQL models do.

\section{Conclusion}

In this work, we identify two issues in current datasets for mapping questions to SQL queries.
First, by analyzing question and query complexity we find that human-written datasets require properties that have not yet been included in large-scale automatically generated query sets.
Second, we show that the generalizability of systems is overstated by the traditional data splits.
In the process we also identify and fix hundreds of mistakes across multiple datasets and homogenize the SQL query structures to enable effective multi-domain experiments.

Our analysis has clear implications for future work. 
Evaluating on multiple datasets is necessary to ensure coverage of the types of questions humans generate. 
Developers of future large-scale datasets should incorporate joins and nesting to create more human-like data.
And new systems should be evaluated on both question- and query- based splits, guiding the development of truly general systems for mapping natural language to structured database queries.

\section*{Acknowledgments}

We would like to thank Laura Wendlandt, Walter Lasecki, and Will Radford for comments on an earlier draft and the anonymous reviewers for their helpful suggestions.
This material is based in part upon work supported by IBM under contract 4915012629. Any opinions, findings, conclusions or recommendations expressed above are those of the authors and do not necessarily reflect the views of IBM.

\bibliographystyle{acl_natbib}
\bibliography{SemanticParsing,DeepLearning}

\begin{thebibliography}{30}
\expandafter\ifx\csname natexlab\endcsname\relax\def\natexlab#1{#1}\fi

\bibitem[{Androutsopoulos et~al.(1995)Androutsopoulos, Ritchie, and
  Thanisch}]{Androutsopoulos1995}
I.~Androutsopoulos, G.~D. Ritchie, and P.~Thanisch. 1995.
\newblock \href {https://doi.org/10.1017/S0269888900005476} {{Natural Language
  Interfaces to Databases - An Introduction}}.
\newblock \emph{Natural Language Engineering}, 1(709):29--81.

\bibitem[{Bahdanau et~al.(2015)Bahdanau, Cho, and Bengio}]{Bahdanau2014}
Dzmitry Bahdanau, Kyunghyun Cho, and Yoshua Bengio. 2015.
\newblock \href {http://arxiv.org/abs/1409.0473} {{Neural machine translation
  by jointly learning to align and translate}}.
\newblock In \emph{Proceedings of the ICLR}, pages 1--15, San Diego,
  California.

\bibitem[{Beltagy et~al.(2014)Beltagy, Erk, and Mooney}]{Beltagy2014}
Islam Beltagy, Katrin Erk, and Raymond Mooney. 2014.
\newblock \href {https://doi.org/10.3115/v1/W14-2402} {{Semantic parsing using
  distributional semantics and probabilistic logic}}.
\newblock \emph{Proceedings of the ACL 2014 Workshop on Semantic Parsing},
  pages 7--11.

\bibitem[{Berant and Liang(2014)}]{Berant2014}
Jonathan Berant and Percy Liang. 2014.
\newblock \href {http://www.aclweb.org/anthology/P14-1133} {{Semantic parsing
  via paraphrasing}}.
\newblock In \emph{Proceedings of the 52nd Annual Meeting of the Association
  for Computational Linguistics}, pages 1415--1425.

\bibitem[{Britz et~al.(2017)Britz, Goldie, Luong, and Le}]{Britz2017}
Denny Britz, Anna Goldie, Minh-Thang Luong, and Quoc Le. 2017.
\newblock \href {https://www.aclweb.org/anthology/D17-1151} {Massive
  exploration of neural machine translation architectures}.
\newblock In \emph{Proceedings of the 2017 Conference on Empirical Methods in
  Natural Language Processing}, pages 1442--1451, Copenhagen, Denmark.
  Association for Computational Linguistics.

\bibitem[{Cai et~al.(2017)Cai, Xu, Yang, Zhang, and Li}]{Cai2017}
Ruichu Cai, Boyan Xu, Xiaoyan Yang, Zhenjie Zhang, and Zijian Li. 2017.
\newblock \href {http://arxiv.org/abs/1711.06061} {{An encoder-decoder
  framework translating natural language to database queries}}.
\newblock \emph{ArXiv e-prints}.

\bibitem[{Dahl et~al.(1994)Dahl, Bates, Brown, Fisher, Hunicke-Smith, Pallett,
  Pao, Rudnicky, and Shriber}]{Dahl1994}
Deborah~A. Dahl, Madeleine Bates, Michael Brown, William Fisher, Kate
  Hunicke-Smith, David Pallett, Christine Pao, Alexander Rudnicky, and
  Elizabeth Shriber. 1994.
\newblock \href {https://doi.org/10.3115/1075812.1075823} {{Expanding the scope
  of the ATIS task: The ATIS-3 corpus}}.
\newblock \emph{Proceedings of the workshop on Human Language Technology},
  pages 43--48.

\bibitem[{Dong and Lapata(2016)}]{Dong2016}
Li~Dong and Mirella Lapata. 2016.
\newblock \href {https://doi.org/10.18653/v1/P16-1004} {{Language to logical
  form with neural attention}}.
\newblock \emph{Proceedings of the 54th Annual Meeting of the Association for
  Computational Linguistics}, 1:33--43.

\bibitem[{Giordani and Moschitti(2012)}]{Giordani2012}
Alessandra Giordani and Alessandro Moschitti. 2012.
\newblock \href {http://www.aclweb.org/anthology/C12-2040} {{Translating
  questions to SQL queries with generative parsers discriminatively reranked}}.
\newblock In \emph{COLING 2012}, pages 401--410.

\bibitem[{Iyer et~al.(2017)Iyer, Konstas, Cheung, Krishnamurthy, and
  Zettlemoyer}]{Iyer2017}
Srinivasan Iyer, Ioannis Konstas, Alvin Cheung, Jayant Krishnamurthy, and Luke
  Zettlemoyer. 2017.
\newblock \href {http://aclweb.org/anthology/P17-1089} {{Learning a neural
  semantic parser from user feedback}}.
\newblock In \emph{Proceedings of the 55th Annual Meeting of the Association
  for Computational Linguistics (Volume 1: Long Papers)}, pages 963----973,
  Vancouver, Canada.

\bibitem[{Jia and Liang(2016)}]{Jia2016}
Robin Jia and Percy Liang. 2016.
\newblock \href {http://www.aclweb.org/anthology/P16-1002} {{Data recombination
  for neural semantic parsing}}.
\newblock In \emph{Proceedings of the 54th Annual Meeting of the Association
  for Computational Linguistics (Volume 1: Long Papers)}, pages 12--22.

\bibitem[{Jiang et~al.(2017)Jiang, Kummerfeld, and Lasecki}]{Jiang2017}
Youxuan Jiang, Jonathan~K. Kummerfeld, and Walter~S. Lasecki. 2017.
\newblock \href {https://doi.org/10.18653/v1/P17-2017} {{Understanding Task
  Design Trade-offs in Crowdsourced Paraphrase Collection}}.
\newblock In \emph{Proceedings of the 55th Annual Meeting of the Association
  for Computational Linguistics (Volume 2: Short Papers)}, pages 103--109,
  Vancouver, Canada.

\bibitem[{Li and Jagadish(2014)}]{Li2014}
Fei Li and H.~V. Jagadish. 2014.
\newblock \href {http://www.vldb.org/pvldb/vol8/p73-li.pdf} {{Constructing an
  interactive natural language interface for relational databases}}.
\newblock In \emph{Proceedings of the VLDB Endowment}, pages 73--84.

\bibitem[{Liang et~al.(2011)Liang, Jordan, and Klein}]{Liang2011}
Percy Liang, Michael~I Jordan, and Dan Klein. 2011.
\newblock \href {http://www.aclweb.org/anthology/P11-1060} {{Learning
  dependency-based compositional semantics}}.
\newblock In \emph{Proceedings of the 49th Annual Meeting of the Association
  for Computational Linguistics: Human Language Technologies - Volume 1}, pages
  590--599, Portland, Oregon.

\bibitem[{Neelakantan et~al.(2017)Neelakantan, Le, Abadi, McCallum, and
  Amodei}]{Neelakantan2017}
Arvind Neelakantan, Quoc~V Le, Mart{\'{i}}n Abadi, Andrew McCallum, and Dario
  Amodei. 2017.
\newblock \href {http://arxiv.org/abs/1611.08945} {{Learning a natural language
  interface with neural programmer}}.
\newblock \emph{Proceedings of the ICLR}, pages 1--10.

\bibitem[{Neubig et~al.(2017)Neubig, Dyer, Goldberg, Matthews, Ammar,
  Anastasopoulos, Ballesteros, Chiang, Clothiaux, Cohn, Duh, Faruqui, Gan,
  Garrette, Ji, Kong, Kuncoro, Kumar, Malaviya, Michel, Oda, Richardson,
  Saphra, Swayamdipta, and Yin}]{dynet}
Graham Neubig, Chris Dyer, Yoav Goldberg, Austin Matthews, Waleed Ammar,
  Antonios Anastasopoulos, Miguel Ballesteros, David Chiang, Daniel Clothiaux,
  Trevor Cohn, Kevin Duh, Manaal Faruqui, Cynthia Gan, Dan Garrette, Yangfeng
  Ji, Lingpeng Kong, Adhiguna Kuncoro, Gaurav Kumar, Chaitanya Malaviya, Paul
  Michel, Yusuke Oda, Matthew Richardson, Naomi Saphra, Swabha Swayamdipta, and
  Pengcheng Yin. 2017.
\newblock \href {https://arxiv.org/abs/1701.03980} {Dynet: The dynamic neural
  network toolkit}.
\newblock \emph{arXiv preprint arXiv:1701.03980}.

\bibitem[{{Pazos Rangel} et~al.(2013){Pazos Rangel}, {Gonz{\'{a}}lez Barbosa},
  {Aguirre Lam}, {Mart{\'{i}}nez Flores}, and {Fraire Huacuja}}]{PazosR.2013}
Rodolfo~A. {Pazos Rangel}, Juan~Javier {Gonz{\'{a}}lez Barbosa}, Marco~Antonio
  {Aguirre Lam}, Jos{\'{e}}~Antonio {Mart{\'{i}}nez Flores}, and
  H{\'{e}}ctor~J. {Fraire Huacuja}. 2013.
\newblock \href
  {https://link.springer.com/chapter/10.1007/978-3-642-33021-6_36}
  {\emph{{Natural language interfaces to databases: An analysis of the state of
  the art}}}. Springer Berlin Heidelberg, Berlin, Heidelberg.

\bibitem[{Poon(2013)}]{Poon2013}
Hoifung Poon. 2013.
\newblock \href {http://www.aclweb.org/anthology/P13-1092} {{Grounded
  unsupervised semantic parsing}}.
\newblock In \emph{Proceedings of the 51st Annual Meeting of the Association
  for Computational Linguistics (Volume 1: Long Papers)}, pages 933--943.

\bibitem[{Popescu et~al.(2004)Popescu, Armanasu, Etzioni, Ko, and
  Yates}]{Popescu2004}
Ana-Maria Popescu, Alex Armanasu, Oren Etzioni, David Ko, and Alexander Yates.
  2004.
\newblock \href {http://www.aclweb.org/anthology/C04-1021} {{Modern natural
  language interfaces to databases: composing statistical parsing with semantic
  tractability}}.
\newblock In \emph{Proceedings of the 20th International Conference on
  Computational Linguistics}, pages 141--147.

\bibitem[{Popescu et~al.(2003)Popescu, Etzioni, and Kautz}]{Popescu2003}
Ana-Maria Popescu, Oren Etzioni, and Henry Kautz. 2003.
\newblock \href {https://doi.org/10.1145/604045.604070} {{Towards a theory of
  natural language interfaces to databases}}.
\newblock \emph{Proceedings of the 8th International Conference on Intelligent
  User Interfaces IUI 03}, pages 149--157.

\bibitem[{Price(1990)}]{Price1990}
Patti~J. Price. 1990.
\newblock \href {https://doi.org/10.3115/116580.116612} {{Evaluation of spoken
  language systems: The ATIS domain}}.
\newblock \emph{Proc. of the Speech and Natural Language Workshop}, pages
  91--95.

\bibitem[{Sutskever et~al.(2014)Sutskever, Vinyals, and Le}]{Sutskever2014}
Ilya Sutskever, Oriol Vinyals, and Quoc~V Le. 2014.
\newblock \href
  {http://papers.nips.cc/paper/5346-sequence-to-sequence-learning-with-neural}
  {{Sequence to sequence learning with neural networks}}.
\newblock \emph{Advances in Neural Information Processing Systems (NIPS)},
  pages 3104--3112.

\bibitem[{Tang and Mooney(2000)}]{Tang2000}
Lappoon~R. Tang and Raymond~J. Mooney. 2000.
\newblock \href {http://www.aclweb.org/anthology/W/W00/W00-1317.pdf}
  {{Automated Construction of Database Interfaces: Integrating Statistical and
  Relational Learning for Semantic Parsing}}.
\newblock \emph{Proceedings of the Joint SIGDAT Conference on Emprical Methods
  in Natural Language Processing and Very Large Corpora}, pages 133--141.

\bibitem[{Yaghmazadeh et~al.(2017)Yaghmazadeh, Wang, Dillig, and
  Dillig}]{Yaghmazadeh2017}
Navid Yaghmazadeh, Yuepeng Wang, Isil Dillig, and Thomas Dillig. 2017.
\newblock \href {http://doi.org/10.1145/3133887} {Sqlizer: Query synthesis from
  natural language}.
\newblock In \emph{International Conference on Object-Oriented Programming,
  Systems, Languages, and Applications, ACM}, pages 63:1--63:26.

\bibitem[{Yih et~al.(2015)Yih, Chang, He, and Gao}]{Yih2015}
Wen-tau Yih, Ming-Wei Chang, Xiaodong He, and Jianfeng Gao. 2015.
\newblock \href {http://www.aclweb.org/anthology/P15-1128} {{Semantic parsing
  via staged query graph generation: Question answering with knowledge base}}.
\newblock In \emph{Proceedings of the 53rd Annual Meeting of the Association
  for Computational Linguistics and the 7th International Joint Conference on
  Natural Language Processing (Volume 1: Long Papers)}, pages 1321--1331,
  Beijing, China.

\bibitem[{Yin et~al.(2016)Yin, Lu, Li, and Kao}]{Yin2016}
Pengcheng Yin, Zhengdong Lu, Hang Li, and Ben Kao. 2016.
\newblock \href {http://arxiv.org/abs/1512.00965} {{Neural Enquirer: Learning
  to query tables in natural language}}.
\newblock \emph{Proceedings of the Twenty-Fifth International Joint Conference
  on Artificial Intelligence (IJCAI-16)}, pages 2308--2314.

\bibitem[{Zelle and Mooney(1996)}]{Zelle1996}
John~M. Zelle and Raymond~J. Mooney. 1996.
\newblock \href {http://www.aaai.org/Papers/AAAI/1996/AAAI96-156.pdf}
  {{Learning to Parse Database queries using inductive logic proramming}}.
\newblock \emph{Learning}, pages 1050--1055.

\bibitem[{Zettlemoyer and Collins(2005)}]{Zettlemoyer2005}
Luke Zettlemoyer and Michael Collins. 2005.
\newblock \href {http://arxiv.org/abs/1207.1420} {{Learning to Map Sentences to
  Logical Form : Structured Classification with Probabilistic Categorial
  Grammars}}.
\newblock \emph{21st Conference on Uncertainty in Artificial Intelligence},
  pages 658--666.

\bibitem[{Zettlemoyer and Collins(2007)}]{Zettlemoyer2007}
Luke Zettlemoyer and Michael Collins. 2007.
\newblock \href {http://aclweb.org/anthology/D/D07/D07-1071.pdf} {Online
  learning of relaxed {CCG} grammars for parsing to logical form}.
\newblock In \emph{Proceedings of the 2007 Joint Conference on Empirical
  Methods in Natural Language Processing and Computational Natural Language
  Learning (EMNLP-CoNLL)}, pages 678--687, Prague, Czech Republic.

\bibitem[{Zhong et~al.(2017)Zhong, Xiong, and Socher}]{Zhong2017}
Victor Zhong, Caiming Xiong, and Richard Socher. 2017.
\newblock \href {http://arxiv.org/abs/1709.00103} {{Seq2SQL: Generating
  Structured Queries from Natural Language using Reinforcement Learning}}.
\newblock \emph{ArXiv e-prints}, pages 1--12.

\end{thebibliography}

\end{document}